\documentclass{article}
\usepackage{ijcai21}

\usepackage{times}

\usepackage{soul}
\usepackage{url}
\usepackage[hidelinks]{hyperref}
\usepackage[utf8]{inputenc}
\usepackage[small]{caption}
\usepackage{graphicx}
\usepackage{amsmath}
\usepackage{booktabs}
\usepackage{algorithm}

\usepackage{amsmath}
\usepackage{amssymb}
\usepackage{color}
\usepackage{epstopdf}
\usepackage{array}
\usepackage{multirow}
\usepackage{amsfonts}
\usepackage{amsthm}
\usepackage{array}
\usepackage{multirow}
\usepackage{amsfonts}
\usepackage{subfigure}
\usepackage{float}
\usepackage{algorithm}
\usepackage{algpseudocode}
\title{Weakly-Supervised Spatio-Temporal Anomaly Detection in Surveillance Video}

\author{
Jie Wu$^{1,3}$\  \and
Wei Zhang$^2$\and
Guanbin Li$^{1}$~\footnote{ Most of the work is done when Jie Wu was a research intern at Baidu. Corresponding author is Guanbin Li. } \and
Wenhao Wu$^2$\and
Xiao Tan$^2$\and \\
Yingying Li$^2$\and
Errui Ding$^2$\And
Liang Lin$^1$\\
\affiliations
$^1$Sun Yat-sen University \\
$^2$Baidu Inc. \\
$^3$ByteDance Inc.\\
\emails
wujie.10@bytedance.com,
\{zhangwei99, wuwenhao01, tanxiao01,liyingying05,dingerrui\}@baidu.com,
liguanbin@mail.sysu.edu.cn,
linliang@ieee.org
}

\begin{document}

\maketitle

\begin{abstract}
    In this paper, we introduce a novel task, referred to as \emph{Weakly-Supervised Spatio-Temporal Anomaly Detection (WSSTAD)} in surveillance video.
    Specifically, given an untrimmed video, WSSTAD aims to localize a spatio-temporal tube (i.e., a sequence of bounding boxes at consecutive times) that encloses the abnormal event, with only coarse video-level annotations as supervision during training.
    To address this challenging task, we propose a dual-branch network which takes as input the proposals with multi-granularities in both spatial-temporal domains.
    Each branch employs a relationship reasoning module to capture the correlation between tubes/videolets, which can provide rich contextual information and complex entity relationships for the concept learning of abnormal behaviors.
    \emph{Mutually-guided Progressive Refinement} framework is set up to employ dual-path mutual guidance in a recurrent manner, iteratively sharing auxiliary supervision information across branches.
    It impels the learned concepts of each branch to serve as a guide for its counterpart, which progressively refines the corresponding branch and the whole framework.
    Furthermore, we contribute two datasets, i.e., \emph{ST-UCF-Crime} and \emph{STRA}, consisting of videos containing spatio-temporal abnormal annotations to serve as the benchmarks for WSSTAD.
    We conduct extensive qualitative and quantitative evaluations to demonstrate the effectiveness of the proposed approach and analyze the key factors that contribute more to handle this task.
\end{abstract}

\section{Introduction}

Anomaly detection in the surveillance video is a fundamental computer vision task and plays a critical role in video structure analysis and potential down-stream applications, e.g., accident forecasting, urban traffic analysis, evidence investigation.
Although it has attracted intense attention in recent years, it remains a very challenging problem due to the serious imbalance between normal and abnormal samples, the lack of fine-grained abnormal labeling data and the ambiguity of the concept of abnormal behavior.
Previous studies \cite{luo2017revisit,hasan2016learning,xu2015learning} generally leverage normal training samples to model abnormal concepts, and identify the distinctive behaviors that deviate from normal patterns as anomalies.
However, these works are not accessible to the abnormal videos, which may incorrectly classify some normal behaviors with abrupt action as abnormal ones, i.e., car acceleration or pedestrian intrusion, resulting in a high false alarm rate.
Recently, some studies \cite{sultani2018real,zhong2019graph,zhang2019temporal,zhu2019motion}  have attempted to introduce videos of abnormal behavior during the training process to better predict which frames or snippets contain abnormal behavior.
However, these works can be regarded as coarse-level detection since they can only predict anomaly in the temporal dimension, but fail to provide more critical region-level abnormal details.
In fact, fine-grained anomaly detection that simultaneously predicts when and where the anomaly happens in the video is more in line with the requirements of the actual application scenario.
For example, during the accident investigation, the traffic polices pay more attention to the objects involved in the accident and their corresponding trajectories, rather than the moment when the abnormality occurred.
Moreover, fine-grained spatio-temporal location of abnormal behaviors can also provide more reliable explanatory guarantees for the anomaly classification.

\begin{figure}[t]
    \centering
    \includegraphics[width=0.99\linewidth]{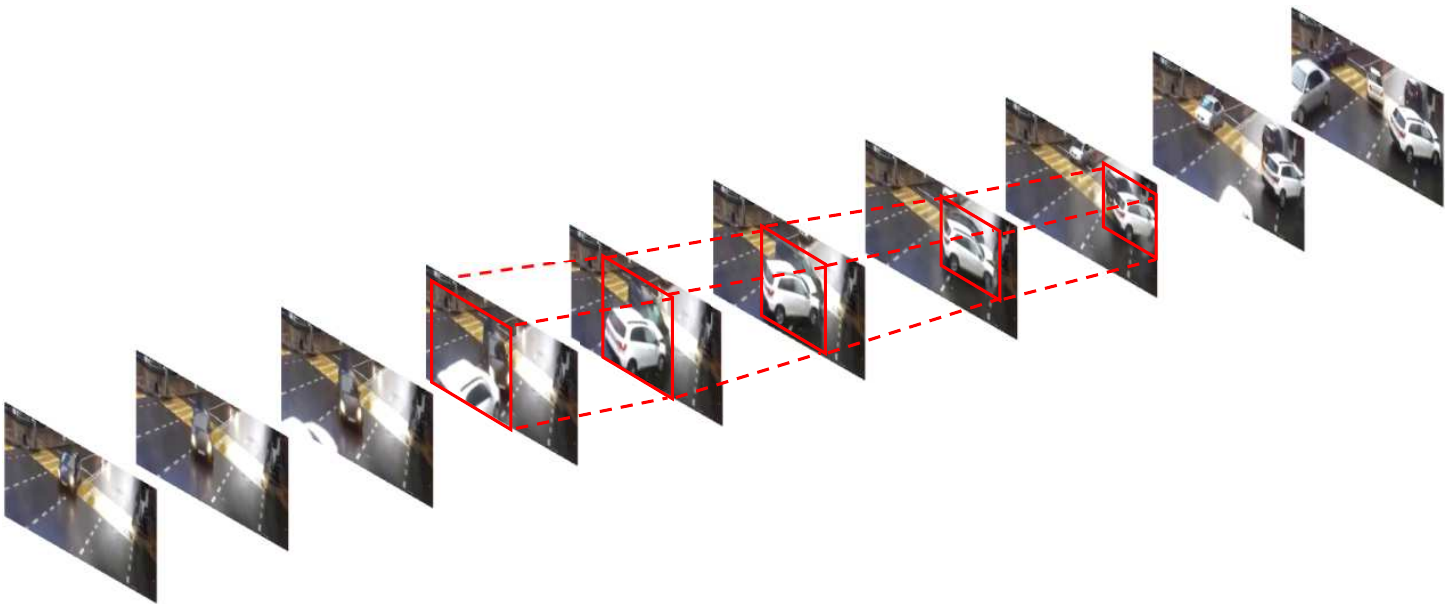}
    \caption{Our proposed WSSTAD task is to localize the spatio-temporal tube of abnormal event (as shown in red) using only video-level label during training.}
    \label{fig:introduction}
\end{figure}

These observations motivate us to introduce a novel task, referred to as \emph{Weakly-Supervised Spatio-Temporal Anomaly Detection (WSSTAD)}. WSSTAD aims to localize a spatio-temporal tube (i.e., a sequence of bounding boxes at consecutive times), which encloses the trajectory of an abnormal event in an untrimmed video.
We follow \cite{sultani2018real} to address this task in a weakly supervised setting that it does not rely on any spatio-temporal annotations during the training process, refer to an example (two cars in the collision) as shown in Figure \ref{fig:introduction}.
Compared with existing anomaly detection problems, our proposed task poses three additional challenges.
1)
The weakly supervised nature of this problem is that both the temporally segment-level labels and spatially region-wise labels are not available during training.
2) This localization task spans spatial and temporal dimensions while the spatial details and temporal correlation can be viewed as cues at different granularity levels. How to leverage such multi-granularity information to jointly facilitate model training remains to be studied.
3) Some anomalies such as ``road accident'' involve the interaction of objects, thus an inherent challenge is to automatically infer the latent relationship between objects in the videos.

In order to tackle this task, we formulate it as a Multiple Instance Learning problem \cite{sultani2018real,chen2019weakly,yamaguchi2017spatio}.
We extract two kinds of tube-level instance proposals and feed them into a tube branch to capture spatial cues.
It is non-trivial to distinguish the anomaly from a single instance, so we propagate information among the instances to make a more comprehensive prediction.
Concretely, each branch employs a relationship modeling module that adopts the multi-head self-attention mechanism to capture the relationships between video objects, and thus incorporate the contextual information and complex entity behavior relationships for anomaly inference.
As each branch helps to capture abnormal abstractions at different granularity level, we can transfer the learned concepts from one branch to the other intuitively.
To this end, we present a novel \emph{Mutually-guided Progressive Refinement} (MGPR) framework, which involves a dual-path mutual guidance mechanism in a recurrent manner to iteratively facilitate the optimization procedure.
Our experiments show that dual-path recurrent guidance coordinates to mutually reinforce two training procedures and boost the performance progressively.


The contributions of this work are summarized into four folds:
1) We present a new task WSSTAD to localize a spatio-temporal tube that semantically corresponds to abnormal event, with no reliance on any spatio-temporal annotations during training.
2) To address this task, MGPR framework is designed to transfer learned abstractions across branches, encouraging mutual guidance and progressive refinement in the whole framework.
3) We contribute two datasets that provide fine-grained tube-level annotations for abnormal videos to serve as the benchmarks.
4) In-depth analyses are conducted to demonstrate the effectiveness of the proposed framework over some competitive methods and discuss the key factors that contribute more to handle this task.

\section{Related work}

\textbf{Anomaly Detection}.
As a long-lasting task in the computer vision field, anomaly detection has been extensively studied for a long time \cite{nallaivarothayan2014mrf,luo2017revisit,hasan2016learning,xu2015learning,sultani2018real,zhong2019graph,zhang2019temporal,zhu2019motion,li2020multi,wu2020modularized,zhao2021good,wu2021box}.
The paradigm for anomaly detection in \cite{sabokrou2018adversarially,sabokrou2017deep} is one-class classification (a.k.a. unary classification), where no anomaly sample is available during training.
Sultani \emph{et al.} \cite{sultani2018real} propose to learn anomaly in the weakly supervised setting that merely resort to video-level labels instead of snippet-level during the training process.
However, these works \cite{sultani2018real,zhong2019graph,zhang2019temporal,zhu2019motion} regard anomaly detection as a frame/segment-level anomaly ranking task and employ AUC metric to evaluate the model performance.
In fact, we believe that abnormal behavior detection should be defined as a fine-grained video understanding task that involves spatio-temporal location, because in actual applications, early warning and subsequent analysis of an abnormal behavior need to include the time and location of specific events.
To this end, we formulate the weakly-supervised spatio-temporal anomaly detection task to predict tube-level anomalies.
compared to unary classification based works \cite{sabokrou2018adversarially,sabokrou2017deep}, our proposed WSSTAD setting is novel and meaningful as it first realizes spatial-temporal anomaly detection with only video-level labels under the binary-classification setting.

\begin{figure*}[t]
    \centering
    \includegraphics[width=0.99\linewidth]{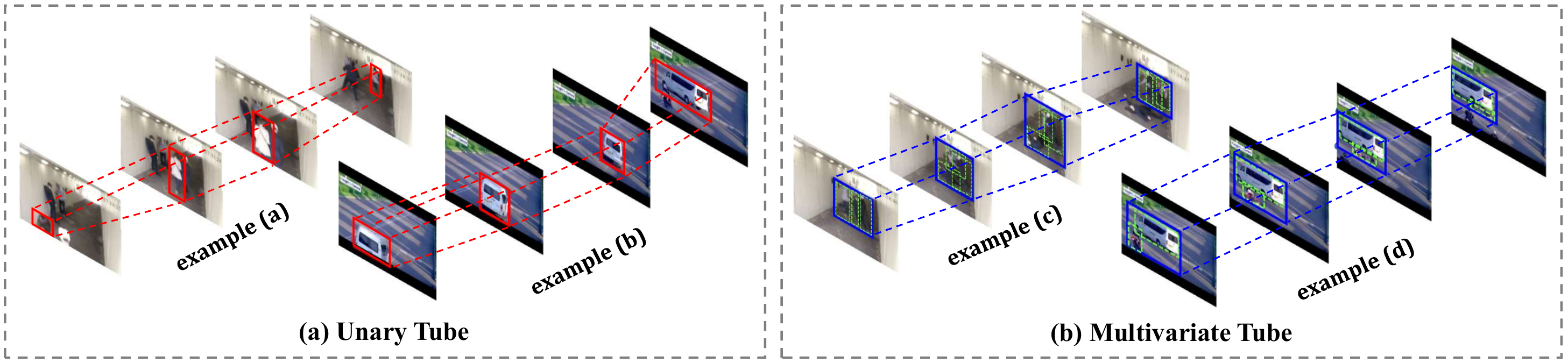}
    \caption{Instances visualizations for ``fighting'' (example (a), (c)) and ``road accident'' (example (b), (d)). In multivariate tubes, each union box (as shown in the blue boxes) may contain several objects (as shown the green boxes).}
    \label{fig:instance}
\end{figure*}

\noindent\textbf{Spatio-Temporal Weakly Supervised Learning}.
Weakly-supervised learning aims at optimizing a model without substantial manual labeling works.
This learning paradigm generally resorts to MIL and applies in many AI tasks such as object detection, captioning and language grounding \cite{wu2020reinforcement}.
Recently, some work have conducted a deep exploration of spatio-temporal weakly supervised learning \cite{chen2019weakly,yamaguchi2017spatio,escorcia2020guess,mettes2018spatio,wu2019multi}, which is studied to localize a tube that corresponds to the task requirement.
Chen \emph{et al.} \cite{chen2019weakly} exploit fine-grained relationships from cross-modal and address the task of grounding natural sentence in the video.
Comparing with weakly supervised s.t. action localization, our task differs significantly and is more challenging due to 1) our proposed task is category agnostic while category info can be used in existing s.t. action localization studies; 2) anomaly essentially differs from human activities; 3) the temporal boundaries of the anomaly (e.g. fight, burglary, accident) are fuzzier than human activities (e.g. surfing, play guitar); 4) the anomalous segment accounts for a much smaller proportion in the entire video (e.g. only 9.28\% for STRA) compared with popular datasets (e.g. UCF101-24 and J-HMDB-21 about 40\%).
%

\section{Methodology}
In this paper, we cast the WSSTAD task as the multiple instance learning (MIL) problem.
We first illustrate the tube-level and videolet-level instances in section 3.1.
Then we describe two relationship modeling module based prediction branches (i.e., tube branch and temporal branch) in section 3.2.
Section 3.3 introduces the proposed mutually-guided progressive refinement framework and the inference procedure is described in Section 3.4.

\subsection{Spatio-Temporal Instance Generation}
We first extract candidate proposals as the instances of MIL.
Conventional anomaly detection works \cite{sultani2018real,zhong2019graph,zhang2019temporal} merely take video snippets as instances, which fail to locate where the anomaly happens spatially.
To this end, we introduce the tube-level instance that links the objects bounding boxes along time into an tube to capture spatio-temporal abnormal cues.
Furthermore, considering that an abnormal event may involve the behavior of a single object or multiple objects, we carefully design two kinds of tube-level instance from different gradations.
As shown in the Figure \ref{fig:instance}, unary tube encloses an individual object and multivariate tube contains multiple intersecting objects.
Multivariate tube is particularly important for capturing the associations between objects which contributes to the localization of abnormal events.
In the experiments, we observe that such multi-gradation instance setting is robust to cope with diverse anomalies. The generation of the tube-level instance is detailed in the supplementary material.
Besides extracting tube-level instances, we also follow \cite{sultani2018real} to utilize temporal correlation to construct videolet-level instances, which is helpful to capture temporal dependencies.

\begin{figure*}[t]
    \centering
    \includegraphics[width=0.99\linewidth]{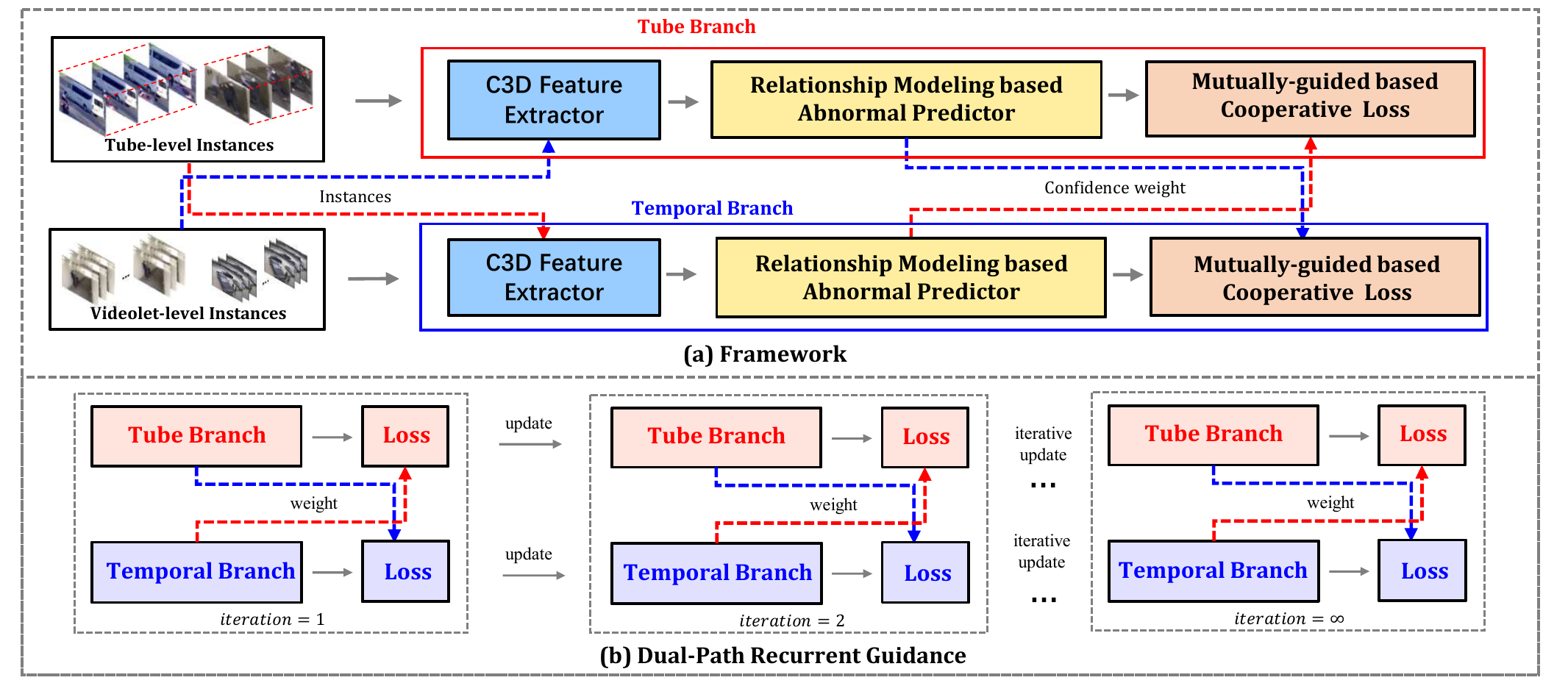}
    \caption{The illustration of \emph{Mutually-guided Progressive Refinement} framework. This framework involves guidance from one branch to the other, promoting mutual enhancement across branches. Dual-path recurrent guidance is performed to progressively facilitate the optimization scheme. The red and blue dashed lines denote the workflow to obtain the confidence weight.}
    \label{fig:framework}
\end{figure*}

\subsection{Relationship Modeling Prediction}

As shown in Figure \ref{fig:framework} (a), 
we accordingly design a dual-branch network architecture, which contains one tube branch that employs tube-level instances to capture spatial cues, and another temporal branch that leverages videolet-level instances for exploiting temporal correlation.

In each branch, the instance features $\mathcal{F}$ are first extracted by a pre-trained C3D model \cite{tran2015learning}.
Previous works \cite{sultani2018real,zhang2019temporal,zhu2019motion} generally fail to take into account the relationship between instances, which ignores the contextual information can provide more reliable cues for comprehensive inference.
To address this issue, we creatively introduce a relationship modeling module that employs the multi-head self-attention \cite{vaswani2017attention} to model the correlations between instances, where the dependency between any two instances is learnable as the attention weight.
The multi-head self-attention \cite{vaswani2017attention} is established on the basis of the scaled dot-product attention $g_a$, which calculates the weights by scaled dot-products of query $\mathrm{\textbf{Q}}$ and keys $\mathrm{\textbf{K}}$. Then $g_a$ computes the output as a weighted sum of values $\mathrm{\textbf{V}}$, formulated as:
\begin{footnotesize}
\begin{equation}
g_{\textit{a}}(\mathrm{\textbf{Q}}, \mathrm{\textbf{K}}, \mathrm{\textbf{V}})=\operatorname{Softmax}\left(\frac{\mathrm{\textbf{Q}} ^{\top} \mathrm{\textbf{K}}}{\sqrt{d_{q}}}\right) \mathrm{\textbf{V}}^{\top},
\end{equation}
\end{footnotesize}
where $d_q$ are the dimension of $\mathrm{\textbf{Q}}$ and the Softmax $(\cdot)$ function is performed along rows.
The multi-head setting deploys $H$ ($H=8$) paralleled attention layers to aggregate information from different representation subspaces and sufficiently decompose the complicated dependencies:
\begin{footnotesize}
\begin{equation}
g_{\textit{ma}}(\mathrm{\bf{Q}}, \mathrm{\bf{K}}, \mathrm{\bf{V}})= \underset{j=1}{\overset{H}\|}h_j^{\top},  h_j =  g_{\textit{a}}(\mathrm{\bf{W^Q_j Q}},\mathrm{\bf{W^K_j K}}, \mathrm{\bf{W^V_j V}}),
\end{equation}
\end{footnotesize}
where $\|$ denotes the concatenation operation on the column dimension, $\mathrm{\bf{W^Q_j}}\in \mathbb{R}^{\frac{d_k}{H} \times d_k}$,$\mathrm{\bf{W^K_j}}\in \mathbb{R}^{\frac{d_k}{H} \times d_k}$, $\mathrm{\bf{W^V_j}}\in \mathbb{R}^{\frac{d_k}{H} \times d_k}$ are linear projection matrices in $j^{th}$ head.
Multi-head self-attention is a special case of the scaled dot-product attention where the $\mathrm{\bf{Q}}$, $\mathrm{\bf{K}}$, $\mathrm{\bf{V}}$ are set to the same video feature matrix $\mathcal{F}=(\mathcal{F}_1$;...;$\mathcal{F}_i$;...;$\mathcal{F}_n) \in \mathbb{R}^{d_k \times n}$. $\mathcal{F}_i \in \mathbb{R}^{d_k}$  denotes the feature of the $i^{th}$ instance and $n$ is the number of instances in the corresponding video.
Then the relationship modeling based self-attentive representation $\tilde{\mathcal{F}}$ = ($\tilde{\mathcal{F}}_1$,.., $\tilde{\mathcal{F}}_n$) is computed via the residual connection:
\begin{equation}
\tilde{\mathcal{F}} =  g_{\textit{ma}}(\mathcal{F}, \mathcal{F}, \mathcal{F}) + \mathcal{F}.
\end{equation}
This residual connection setting maintains original concepts and aggregate contextual information from other instances.
The obtained $\tilde{\mathcal{F}}_i$ is then fed into a three-layer fully connected neural network (abnormal score predictor) \cite{sultani2018real} to predict the abnormal score.

\subsection{Mutually-guided Progressive Refinement}
\noindent\textbf{Multiple Instance Learning}.
Following the setting of MIL, we divide the instances belonging to the anomalous video into the positive bag and the instances for normal videos are putted into the negative bag.
The optimization goal of MIL is to push abnormal instances apart from normal instances in terms of the abnormal score.
Considering that there may be only one abnormal instance in a positive bag, we take the instance that obtains the max abnormal score over all instances in each bag (named as \emph{max instance}) to compute the loss function.
Additionally, we first explain some abbreviations used in the following section.
$\mathcal{V}^{a}$ and $\mathcal{V}^{n}$ denote the videolet-level instances in the anomalous and normal video.
For the tube instances, the sequences of RGB images can be obtained from the inside of tubes (region-level) or from the entire image (image-level).
So we use $\mathcal{T}^{a,r}$, $\mathcal{T}^{a,g}$ to respectively denote the region-level based tube instances and image-level based tube instances in the anomalous video.
$\mathcal{T}^{n,r}$ is the region-level based tube instances in the normal video.
Furthermore, we use $p_{t}$ and $p_{v}$ to represent the scores predicted by the tube and temporal branch, respectively.

\noindent\textbf{Dual-Path Mutual Guidance}.
Our ranking loss in MIL builds upon the existing observations that the learned concepts from the tube and temporal branch are often complementary to each other in action recognition and localization task \cite{saha2016deep,simonyan2014two}.
In the tube-temporal dual branches, we can transfer the learned concepts from one branch to the other branch intuitively, and the concepts can be leveraged as auxiliary supervision to encourage the other branch to learn anomaly from other granularity level.
These observations motivate us to propose the \emph{Mutually-guided Progressive Refinement} framework, which exploits dual-path mutual guidance across branches and refines tube and temporal branches in a progressive manner.
As shown in the Figure \ref{fig:framework}, a novel mutually-guided ranking loss is designed to leverage the feedback of each branch to serve as a guidance for its counterpart, making both branches mutually guided.
Specifically, we feed the max instance of the tube branch into the temporal branch and outputs an abnormal score, which is treated as a confidence weight of the ranking loss that guides the training of the tube branch:
\begin{footnotesize}
\begin{equation}
\begin{split}
\mathcal{L}_{\textit{MG-Rank}}^{\textit{tube}} = \max(0, p_{v}(\mathcal{T}^{a,g}_{m}) \times (1- p_{t}(\mathcal{T}^{a,r}_{m}) +   p_{t}(\mathcal{T}^{n,r}_{m}))),
\end{split}
\end{equation}
\end{footnotesize}
where $\mathcal{T}^{a, r}_{m}$, $\mathcal{T}^{a, g}_{m}$, $\mathcal{T}^{n, r}_{m}$ denotes the corresponding max instance obtained in the tube branch.
We feed $\mathcal{T}^{a, g}_{m}$ instead of $\mathcal{T}^{a, r}_{m}$ into the temporal branch, which ensures the input adapts to the characteristics of the temporal branch.
From the equation we can see that the confidence of the ranking loss is adaptively adjusted by the abnormal concepts already learned by the other branch, instead of fixing it to 1 gruffly in the conventional MIL paradigm \cite{sultani2018real}.
If the temporal branch holds that the max instance is less related to the anomaly (the returned score is lower), it will penalize the ranking loss of the tube branch.
Similarly, the temporal branch is optimized by the guidance from the tube branch:
\begin{footnotesize}
\begin{equation}
\begin{split}
\mathcal{L}_{\textit{MG-Rank}}^{\textit{tem}} = \max(0, p_{t}(\mathcal{V}^{a}_{m})\times (1- p_{v}(\mathcal{V}^{a}_{m}) + p_{v}(\mathcal{V}^{n}_{m}))),
\end{split}
\end{equation}
\end{footnotesize}
where $\mathcal{V}^{a}_{m}$, $\mathcal{V}^{n}_{m}$ denotes the max instance in temporal branch.

\begin{table*}[t]
    \footnotesize
    \centering
    \caption{The performance (in \%) with state-of-the-art methods. The top entry of all the methods except the upper bound is highlighted.}
     \begin{tabular}{m{4.1cm}<{\centering}|m{1.0cm}<{\centering}|m{1.2cm}<{\centering}|m{1.2cm}<{\centering}
     |m{1.0cm}<{\centering}|m{1.0cm}<{\centering}|m{1.2cm}<{\centering}|m{1.2cm}<{\centering}|m{1.0cm}<{\centering}}
    \toprule
    \multicolumn{1}{c|}{} & \multicolumn{4}{c|}{ST-UCF-Crime}  & \multicolumn{4}{c}{STRA}\\ \hline
    Baseline &VAUC  & IoU@0.3 & IoU@0.1 & MIoU &VAUC & IoU@0.2 & IoU@0.1 & MIoU \\ \hline
    Random &44.69	&2.52 & 10.08& 2.64&43.80&3.17&4.76&1.33\\
    Upper Bound &100.00	&31.93& 78.99&25.63	&100.00&37.40&63.56&18.62\\ \hline
    DMRM \cite{sultani2018real} &84.28	&9.24 &21.01 &6.47	&89.37&7.94&9.52&4.14 \\
    GCLNC \cite{zhong2019graph} &85.53	&10.92& \textbf{25.21}& 8.63 & 91.60 & 12.70&17.46&5.41\\ \hline
    STIL \cite{mettes2018spatio} &85.16&9.24&21.84&8.07	&91.23&12.70&15.87&5.19 \\
    ASA \cite{escorcia2020guess} &86.33	&\textbf{11.76}& 23.52& 8.41 & 92.17 & 14.29&18.49&6.54\\ \hline
    Ours &\textbf{87.65}	&\textbf{11.76} &24.37 & \textbf{8.98} & \textbf{92.88} &\textbf{15.87}&\textbf{20.63}&\textbf{7.23}  \\
    \bottomrule
    \end{tabular}
    \label{table:result1}
     \end{table*}

In order to obtain the weight from the tube branch, we simply regard $\mathcal{V}^{a}_{m}$ as tubes and feed 
$\mathcal{V}^{a}_{m}$ into the tube branch. 
By introducing dual-path mutual guidance, the diversity of abnormal abstraction from one branch will be enhanced by using the complementary granularity concepts from the others, which intrinsically improves the anomaly learning ability of both branches.
In each training iteration, we first freeze the tube branch and provide loss weights to optimize the temporal branch. Then we switch to training the tube branch and obtain weights from the temporal branch.
This alternate optimization procedure is repeated iteratively during training:
\begin{equation}
 \mathcal{L}_{\textit{MG-Rank}} =  \mathcal{\psi} \times \mathcal{L}_{\textit{MG-Rank}}^{\textit{tube}} + (1-\mathcal{\psi}) \times \mathcal{L}_{\textit{MG-Rank}}^{\textit{tem}},
\end{equation}
where $\psi$ is a binary variable indicating the selection of the training branch.
With the training going on, as depicted in Figure \ref{fig:framework} (b), the dual-path mutual guidance can be regraded as a recurrent guidance scheme.
Concretely, the outputs from the tube branch gradually provide accurate guidance for the temporal branch. Meanwhile, with the increasingly informative guidance from the temporal branch, tube branch is refined in the loop.
This recurrent guidance scheme contributes to iteratively sharing auxiliary supervision across branches and progressively enhance each branch and the whole framework.

\noindent\textbf{Cross Entropy Loss}. During training, we enforce that the predictor can produce high scores for the abnormal instances, while the normal instances get low scores.
Hence the cross-entropy loss is adopted to encourage the score of max instances aligning to the video-level label. Formally, the score of the max instance in the anomalous/normal should get close to target $1$/$0$:
\begin{equation}
\begin{split}
\mathcal{L}_{\textit{CE}} = -[ \log (p_{t}(\mathcal{T}^{a,r}_{m})) +  \log (1- p_{t}(\mathcal{T}^{n,r}_{m}))]\\
-[ \log (p_{v}(\mathcal{V}^{a}_{m})) +  \log (1- p_{v}(\mathcal{V}^{n}_{m}))].
\end{split}
\end{equation}
Cross entropy loss also ensures that the model does not optimize $\mathcal{L}_{\textit{Rank}}$ by predicting all the scores as zero.
To sum up, mutually-guided ranking loss and cross-entropy loss are combined to optimize the proposed framework jointly:
\begin{equation}
 \mathcal{L} = \mathcal{L}_{\textit{MG-Rank}} + \mathcal{L}_{\textit{CE}}.
\end{equation}

\subsection{Inference}
In the inference stage, we treat the tube-level instances as the hypothetical instances.
Considering that a tube-level instance may have abnormal and normal events, we evenly divide each instance into $M$ hypothetical tube instances $\{\mathcal{T}_i\}_{i=1}^{M}$.
Then we input the region-level instances into the tube branch to predict an abnormal score $p_{t}(\mathcal{T}^{r}_i)$.
Subsequently, the global-level instances are fed into the temporal branch to get the corresponding score $p_{v}(\mathcal{T}^{g}_i)$.
Finally, we calculate the prediction scores and retrieve the top-1 tube $\mathcal{T}_{\textit{pred}}$ via employing an aggregate function that averages scores from two branches:
\begin{equation}
\mathcal{T}_{\textit{pred}} = \mathop{\arg\max}_{\mathcal{T}_i}\frac{p_{t}(\mathcal{T}^{r}_i) + p_{v}(\mathcal{T}^{g}_i)}{2} .
\end{equation}


\section{Spatio-Temporal Anomaly Dataset}

A major challenge for the proposed WSSTAD task is the lack of appropriate datasets.
UCF-Crime and other anomaly detection datasets \cite{luo2017revisit,li2013anomaly} do not provide spatio-temporal annotations for the ground truth instances, which are necessary for evaluation.
To this end, we build a new dataset (denoted as ST-UCF-Crime) that annotates spatio-temporal bounding boxes for abnormal events in UCF-Crime \cite{sultani2018real}, which contains anomaly videos of diverse categories in complicated surveillance scenarios.
Furthermore, we contribute a new dataset, namely Spatio-Temporal Road Accident (abbreviated as STRA) that consists of various road accidents videos, such as motorcycles crash into cars, cars crash into people, etc.. STRA contributes to fine-grained anomaly detection in actual traffic accident scenarios and promoting the development of intelligent transportation. We provide more details in the appendix.


\section{Experiments}
\subsection{Experimental Setup}
\textbf{Implementation Details}.
We feed each 16-frame image sequence within the instance into C3D \cite{tran2015learning} and extract the features from the \emph{fc6} layer. Then we take a mean pooling layer to average these features and obtain the instance feature.
We randomly choose 30 positive and 30 negative bags to construct a mini-batch, and the number of instances per bag is limited to 200.
The total loss is optimized via Adam optimizer with the learning rate of 0.0005.

\noindent \textbf{Evaluation Metric}.
1) Video-level AUC. The abnormal score of top-$1$ tube $\mathcal{T}_{\textit{pred}}$ is regarded as the abnormal score of the whole video.
We use abnormal score to perform receiver operating characteristic curve and area under the curve is viewed as video-level AUC (VAUC). VAUC is adopted to evaluate the model's ability to distinguish between normal and abnormal video.
2) We utilize spatio-temporal localization score \cite{yamaguchi2017spatio} to measure the overlap of
$\mathcal{T}_{\textit{pred}}$ and ground-truth tube $\mathcal{T}_{\textit{gt}}$:
\begin{equation}
S_{\textit{loc}}(\mathcal{T}_{\textit{gt}}, \mathcal{T}_{\textit{pred}}) = \frac{\sum_{f \in \Gamma} \textit{IoU}(\mathcal{T}^f_{\textit{gt}}, \mathcal{T}^f_{\textit{pred}})}{|\Gamma|},
\end{equation}
where $\textit{IoU}$ denotes the intersection over union between those two bounding boxes.
$\Gamma$ is the intersection of two set of frames $f$. The first set contains the frames have bounding boxes provided by detector. The second set includes the frames in which $\mathcal{T}_{\textit{gt}}$ or $\mathcal{T}_{\textit{pred}}$ has any bounding box.
We define two metrics to evaluate spatio-temporal localization accuracy for the abnormal testing videos. ``IoU@$\epsilon$'' means the percentage of the videos that have $S_{\textit{loc}}$ larger than $\epsilon$.
``MIoU'' denotes the average IoU for all abnormal testing videos.

\subsection{Comparison with the State-of-the-art}
We compare our approach with some state-of-the-art weakly-supervised anomaly detection approaches, DMRM \cite{sultani2018real} and GCLNC \cite{zhong2019graph} in Table \ref{table:result1}.
We also compare with some weakly-supervised spatio-temporal action localization approaches, STLA\cite{mettes2018spatio} and ASA \cite{escorcia2020guess}.
We additionally show the performance of randomly selecting a candidate tube and the upper bound performance of choosing the tube of the largest overlap with the ground-truth.

As shown in the Table \ref{table:result1}, our approach significantly exceeds the performance of random selection in VAUC or MIoU. Furthermore, our approach outperforms \cite{sultani2018real,zhang2019temporal,zhong2019graph,mettes2018spatio,escorcia2020guess} and achieves best performance on both datasets.
From video-level metric, our method improves VAUC by 1.32\% and 0.71\% compared with the previous best \cite{escorcia2020guess} on two datasets, respectively.
It reveals that our approach helps to better determine whether the video is abnormal.
For spatio-temporal localization accuracy, the MIoU of our approach achieves 7.23\% (8.98\%) in STRA (ST-UCF-Crime), obtaining comparative enhancement over \cite{escorcia2020guess} by 10.55\% (7.02\%).
In the contrast experiment with other methods \cite{sultani2018real,zhang2019temporal,zhong2019graph}, our approach shows a more significant performance advantage on STRA than ST-UCF-Crime.
It may be due to the fact that STRA is more demanding in understanding the relationship between objects, while other methods do not pay much attention to the relationship between objects with interaction.

  \begin{table}[t]
\footnotesize
\centering
\caption{The performance of different learning branches.}
 \begin{tabular}{m{1.1cm}<{\centering}|m{0.8cm}<{\centering}|m{1.0cm}<{\centering}|m{1.0cm}<{\centering}|m{1.0cm}<{\centering}|m{1.0cm}<{\centering}}
\toprule
\multicolumn{2}{c|}{}  & \multicolumn{2}{c|}{ST-UCF-Crime}  & \multicolumn{2}{c}{STRA}\\ \hline
Temporal & Tube &VAUC  & MIoU &VAUC & MIoU \\ \hline
\checkmark&&86.41& 8.26& 90.04 &5.39 \\
&\checkmark&87.15&8.43	&92.40 &6.17 \\
\checkmark&\checkmark&\textbf{87.65} & \textbf{8.98} & \textbf{92.88}&\textbf{7.23}  \\
\bottomrule
\end{tabular}
\label{table:result2}
 \end{table}

\subsection{Ablation Study}

\noindent \textbf{The Effectiveness of Dual Branches}.
To validate the significance of dual branches for modeling abnormal concepts, we design three variants that use a separate branch from our framework to predict anomaly and the results are reported in Table \ref{table:result2}.
``Temporal'' is the temporal branch and ``Tube'' denotes the tube branch that utilizes region-level RGB images.
As shown in Table \ref{table:result2}, the performance of ``Tube'' is superior to ``Temporal'', which shows that tube-level abnormal cues are more effective than the temporal correlation of the abnormal segments.
Furthermore, getting rid of temporal or tube branch will cause performance degradation.
It indicates that exploiting the complementary of abnormal abstractions from dual branches contributes to further boosting the performance.

 \begin{table}[t]
\footnotesize
\centering
\caption{The performance of different loss functions.}
 \begin{tabular}{m{0.7cm}<{\centering}|m{1.0cm}<{\centering}|m{0.7cm}<{\centering}|m{0.9cm}<{\centering}
 |m{0.9cm}<{\centering}|m{0.9cm}<{\centering}|m{0.9cm}<{\centering}}
\toprule
\multicolumn{3}{c|}{}  & \multicolumn{2}{c|}{ST-UCF-Crime}  & \multicolumn{2}{c}{STRA}\\ \hline
$\mathcal{L}_{\textit{Rank}}$ & $\mathcal{L}_{\textit{MG-Rank}}$& $\mathcal{L}_{\textit{CE}}$ &VAUC   & MIoU &VAUC & MIoU \\ \hline
\checkmark &&	&87.31&8.22	&92.17 &6.61 \\
 &\checkmark& &76.83  & 8.10& 92.26 &6.53 \\
 & &	\checkmark&87.57&8.39&91.94 &5.88 \\
 \checkmark& & \checkmark&87.60	  & 8.55 & 92.78 &6.77  \\
 & \checkmark & \checkmark &\textbf{87.65}  & \textbf{8.98} & \textbf{92.88} &\textbf{7.23}  \\
\bottomrule
\end{tabular}
\label{table:result3}
 \end{table}

  \begin{table}[t]
\footnotesize
\centering
\caption{The performance of different interaction methods.}
 \begin{tabular}{m{0.7cm}<{\centering}|m{0.7cm}<{\centering}|m{0.7cm}<{\centering}|m{0.9cm}<{\centering}
|m{0.9cm}<{\centering}|m{0.9cm}<{\centering}|m{0.9cm}<{\centering}}
\toprule
\multicolumn{3}{c|}{}  & \multicolumn{2}{c|}{ST-UCF-Crime}  & \multicolumn{2}{c}{STRA}\\ \hline
MVT & RMM &GCN &VAUC  & MIoU &VAUC  & MIoU \\ \hline
 &\checkmark& &84.08&8.29&91.42&5.66 \\
 \checkmark& & &86.59	 & 8.17& 92.08 &6.55 \\
\checkmark & & \checkmark	&87.42&8.32	&92.33 &6.44 \\
\checkmark &\checkmark & &\textbf{87.65}  & \textbf{8.98} & \textbf{92.88} &\textbf{7.23}  \\
\bottomrule
\end{tabular}
\label{table:result4}
 \end{table}

\noindent \textbf{The Effectiveness of Loss Function}.
To investigate the setting of the loss function, we further design four baselines with different loss functions and summarize the results in Table \ref{table:result3}.
As shown in Table \ref{table:result3}, the baseline with the ranking loss $\mathcal{L}_{\textit{Rank}}$ \cite{sultani2018real} gets an unimpressive result.
Employing mutual guidance to the ranking loss further declines the performance. It is beacuse the model learns to reduce the loss function via predicting all scores to approach 0 trickly.
$\mathcal{L}_{\textit{CE}}$ helps to alleviate this issue and cooperate with $\mathcal{L}_{\textit{MG-Rank}}$ to obtain promising results. We can see that our approach with these two losses can obtain 8.98 (7.23) in ST-UCF-Crime (STRA) in terms of MIOU, which performs best among the baselines in Table \ref{table:result3}.
The variant discards the mutual guidance mechanism based on our approach suffers from performance degradation.
The above observations reveal that our approach helps to take full advantage of auxiliary supervision information to achieve accurate anomaly inference.

\noindent \textbf{The Effectiveness of Interaction Methods}.
It may be difficult to detect anomaly from a single object, hence the interaction between objects has an important impact on learning anomalies cues. To explore different interaction methods in our model, we design several variants and present the results in Table \ref{table:result4}.
We observe that VAUC declines from 87.65\% to 84.08\% in ST-UCF-Crime without multivariate tube (MVT).
A similar performance degradation can be observed in the absence of relationship modeling module (RMM). It reveals the effectiveness of the MVT and the RMM.
In fact, MVT and RMM can be viewed as the interaction of the objects at different levels.
MVT associates the intersecting objects to construct the tube, while RMM establishes the relationship between different objects among the tubes. These two levels of interaction contribute to fully mining and propagating the relationship between video objects, encouraging the model to inferring the abnormal event accurately.
Furthermore, we try to replace RMM with graph convolutional network (GCN) designed in \cite{zhong2019graph}.
From Table \ref{table:result4}, we find that although the baseline with GCN can achieve promising results, our model with RMM gains 0.66\% and 0.79\% improvement w.r.t. MIoU in the two datasets respectively.
It may be due to the fact that the adjacency matrix of GCN in is prior and hand-crafted, while the corresponding relationship weights in our approach are obtained via feature learning.

\begin{table}[t]
    \footnotesize
    \centering
    \caption{Video-level AUC for different methods.}
     \begin{tabular}{m{4.1cm}<{\centering}|m{2.0cm}<{\centering}|m{1.0cm}<{\centering}}
    \toprule
    Baseline &   ST-UCF-Crime & STRA \\ \hline
    ALOCC \cite{sabokrou2018adversarially} &68.34&71.96 \\
     Ours &87.65 &92.88 \\
    \bottomrule
    \end{tabular}
    \label{table:result5}
     \end{table}

\subsection{Compared with one-class classification method}
We follow \cite{sabokrou2018adversarially} to feed the normal samples for training and obtain the score from the outputs of $\mathcal{R}$+$\mathcal{D}$ network to compute VAUC. Table \ref{table:result5} reveals that our method can obtain promising results.

\section{Conclusions}
This work explores WSSTAD, a novel task that aims to localize a tube that semantically corresponds to the abnormal event.
To handle this task, we design two prediction branches based on the relationship modeling module to exploit multi-granularity abnormal concepts and establish instances relationship for comprehensive inference.
Mutually-guided Progressive Refinement framework is designed to transfer the learned abnormal concepts of each branch to the other, employing a dual-path recurrent guidance scheme to facilitate mutual guidance across branches and refine the optimization process progressively.
To evaluate this task, we contribute two datasets and conduct extensive experiments to analyze the key factors that contribute more to this task.

\bibliographystyle{named}
\bibliography{WSSTAD}

\begin{thebibliography}{}

\bibitem[\protect\citeauthoryear{Chan \bgroup \em et al.\egroup
  }{2016}]{chan2016anticipating}
Fu-Hsiang Chan, Yu-Ting Chen, Yu~Xiang, and Min Sun.
\newblock Anticipating accidents in dashcam videos.
\newblock pages 136--153. Springer, 2016.

\bibitem[\protect\citeauthoryear{Chen \bgroup \em et al.\egroup
  }{2019}]{chen2019weakly}
Zhenfang Chen, Lin Ma, Wenhan Luo, and Kwan-Yee~Kenneth Wong.
\newblock Weakly-supervised spatio-temporally grounding natural sentence in
  video.
\newblock In {\em Proceedings of the 57th Annual Meeting of the Association for
  Computational Linguistics}, pages 1884--1894, 2019.

\bibitem[\protect\citeauthoryear{Escorcia \bgroup \em et al.\egroup
  }{2020}]{escorcia2020guess}
Victor Escorcia, Cuong~Duc Dao, Mihir Jain, Bernard Ghanem, and Cees G~M Snoek.
\newblock Guess where? actor-supervision for spatiotemporal action
  localization.
\newblock {\em Computer Vision and Image Understanding}, 2020.

\bibitem[\protect\citeauthoryear{Hasan \bgroup \em et al.\egroup
  }{2016}]{hasan2016learning}
Mahmudul Hasan, Jonghyun Choi, Jan Neumann, Amit~K Roy-Chowdhury, and Larry~S
  Davis.
\newblock Learning temporal regularity in video sequences.
\newblock In {\em Proc. CVPR}, pages 733--742, 2016.

\bibitem[\protect\citeauthoryear{Li \bgroup \em et al.\egroup
  }{2013}]{li2013anomaly}
Weixin Li, Vijay Mahadevan, and Nuno Vasconcelos.
\newblock Anomaly detection and localization in crowded scenes.
\newblock {\em IEEE Trans. on Pattern Analysis and Machine Intelligence},
  36(1):18--32, 2013.

\bibitem[\protect\citeauthoryear{Li \bgroup \em et al.\egroup
  }{2020}]{li2020multi}
Yingying Li, Jie Wu, Xue Bai, Xipeng Yang, Xiao Tan, Guanbin Li, Shilei Wen,
  Hongwu Zhang, and Errui Ding.
\newblock Multi-granularity tracking with modularlized components for
  unsupervised vehicles anomaly detection.
\newblock In {\em Proc. CVPRW}, pages 586--587, 2020.

\bibitem[\protect\citeauthoryear{Luo \bgroup \em et al.\egroup
  }{2017}]{luo2017revisit}
Weixin Luo, Wen Liu, and Shenghua Gao.
\newblock A revisit of sparse coding based anomaly detection in stacked rnn
  framework.
\newblock In {\em Proc. ICCV}, pages 341--349, 2017.

\bibitem[\protect\citeauthoryear{Mettes and Snoek}{2018}]{mettes2018spatio}
Pascal Mettes and Cees~GM Snoek.
\newblock Spatio-temporal instance learning: Action tubes from class
  supervision.
\newblock {\em arXiv preprint arXiv:1807.02800}, 2018.

\bibitem[\protect\citeauthoryear{Nallaivarothayan \bgroup \em et al.\egroup
  }{2014}]{nallaivarothayan2014mrf}
Hajananth Nallaivarothayan, Clinton Fookes, Simon Denman, and Sridha Sridharan.
\newblock An mrf based abnormal event detection approach using motion and
  appearance features.
\newblock In {\em IEEE International Conference on Advanced Video and Signal
  Based Surveillance (AVSS)}, pages 343--348. IEEE, 2014.

\bibitem[\protect\citeauthoryear{Ren \bgroup \em et al.\egroup
  }{2015}]{ren2015faster}
Shaoqing Ren, Kaiming He, Ross Girshick, and Jian Sun.
\newblock Faster r-cnn: Towards real-time object detection with region proposal
  networks.
\newblock In {\em Neurips}, pages 91--99, 2015.

\bibitem[\protect\citeauthoryear{Sabokrou \bgroup \em et al.\egroup
  }{2017}]{sabokrou2017deep}
Mohammad Sabokrou, Mohsen Fayyaz, Mahmood Fathy, and Reinhard Klette.
\newblock Deep-cascade: Cascading 3d deep neural networks for fast anomaly
  detection and localization in crowded scenes.
\newblock {\em IEEE Trans. on Image Processing}, 26(4):1992--2004, 2017.

\bibitem[\protect\citeauthoryear{Sabokrou \bgroup \em et al.\egroup
  }{2018}]{sabokrou2018adversarially}
Mohammad Sabokrou, Mohammad Khalooei, Mahmood Fathy, and Ehsan Adeli.
\newblock Adversarially learned one-class classifier for novelty detection.
\newblock In {\em Proc. CVPR}, pages 3379--3388, 2018.

\bibitem[\protect\citeauthoryear{Saha \bgroup \em et al.\egroup
  }{2016}]{saha2016deep}
Suman Saha, Gurkirt Singh, Michael Sapienza, Philip~HS Torr, and Fabio
  Cuzzolin.
\newblock Deep learning for detecting multiple space-time action tubes in
  videos.
\newblock {\em arXiv preprint arXiv:1608.01529}, 2016.

\bibitem[\protect\citeauthoryear{Simonyan and
  Zisserman}{2014}]{simonyan2014two}
Karen Simonyan and Andrew Zisserman.
\newblock Two-stream convolutional networks for action recognition in videos.
\newblock In {\em Neurips}, pages 568--576, 2014.

\bibitem[\protect\citeauthoryear{Sultani \bgroup \em et al.\egroup
  }{2018}]{sultani2018real}
Waqas Sultani, Chen Chen, Mubarak Shah, and Waqas Sultani.
\newblock Real-world anomaly detection in surveillance videos.
\newblock In {\em Proc. CVPR}, pages 6479--6488, 2018.

\bibitem[\protect\citeauthoryear{Tran \bgroup \em et al.\egroup
  }{2015}]{tran2015learning}
Du~Tran, Lubomir Bourdev, Rob Fergus, Lorenzo Torresani, and Manohar Paluri.
\newblock Learning spatiotemporal features with 3d convolutional networks.
\newblock In {\em Proc. ICCV}, pages 4489--4497, 2015.

\bibitem[\protect\citeauthoryear{Vaswani \bgroup \em et al.\egroup
  }{2017}]{vaswani2017attention}
Ashish Vaswani, Noam Shazeer, Niki Parmar, Jakob Uszkoreit, Llion Jones,
  Aidan~N Gomez, {\L}ukasz Kaiser, and Illia Polosukhin.
\newblock Attention is all you need.
\newblock In {\em Neurips}, pages 5998--6008, 2017.

\bibitem[\protect\citeauthoryear{Wu \bgroup \em et al.\egroup
  }{2019}]{wu2019multi}
Wenhao Wu, Dongliang He, Xiao Tan, Shifeng Chen, and Shilei Wen.
\newblock Multi-agent reinforcement learning based frame sampling for effective
  untrimmed video recognition.
\newblock In {\em Proc. ICCV}, 2019.

\bibitem[\protect\citeauthoryear{Wu \bgroup \em et al.\egroup
  }{2020a}]{wu2020reinforcement}
Jie Wu, Guanbin Li, Xiaoguang Han, and Liang Lin.
\newblock Reinforcement learning for weakly supervised temporal grounding of
  natural language in untrimmed videos.
\newblock In {\em Proc. ACMMM}, pages 1283--1291, 2020.

\bibitem[\protect\citeauthoryear{Wu \bgroup \em et al.\egroup
  }{2020b}]{wu2020modularized}
Jie Wu, Yingying Li, Wei Zhang, Yi~Wu, Xiao Tan, Hongwu Zhang, Shilei Wen,
  Errui Ding, and Guanbin Li.
\newblock Modularized framework with category-sensitive abnormal filter for
  city anomaly detection.
\newblock In {\em Proc. ACMMM}, pages 4669--4673, 2020.

\bibitem[\protect\citeauthoryear{Wu \bgroup \em et al.\egroup
  }{2021}]{wu2021box}
Jie Wu, Xionghui Wang, Xuefeng Xiao, and Yitong Wang.
\newblock Box-level tube tracking and refinement for vehicles anomaly
  detection.
\newblock In {\em Proc. CVPRW}, 2021.

\bibitem[\protect\citeauthoryear{Xu \bgroup \em et al.\egroup
  }{2015}]{xu2015learning}
Dan Xu, Elisa Ricci, Yan Yan, Jingkuan Song, and Nicu Sebe.
\newblock Learning deep representations of appearance and motion for anomalous
  event detection.
\newblock {\em arXiv preprint arXiv:1510.01553}, 2015.

\bibitem[\protect\citeauthoryear{Yamaguchi \bgroup \em et al.\egroup
  }{2017}]{yamaguchi2017spatio}
Masataka Yamaguchi, Kuniaki Saito, Yoshitaka Ushiku, and Tatsuya Harada.
\newblock Spatio-temporal person retrieval via natural language queries.
\newblock In {\em Proc. ICCV}, pages 1453--1462, 2017.

\bibitem[\protect\citeauthoryear{Zhang \bgroup \em et al.\egroup
  }{2019}]{zhang2019temporal}
Jiangong Zhang, Laiyun Qing, Jun Miao, and Jiangong Zhang.
\newblock Temporal convolutional network with complementary inner bag loss for
  weakly supervised anomaly detection.
\newblock In {\em Proc. ICIP}, pages 4030--4034. IEEE, 2019.

\bibitem[\protect\citeauthoryear{Zhao \bgroup \em et al.\egroup
  }{2021}]{zhao2021good}
Yuxiang Zhao, Wenhao Wu, Yue He, Yingying Li, Xiao Tan, and Shifeng Chen.
\newblock Good practices and a strong baseline for traffic anomaly detection.
\newblock In {\em Proc. CVPRW}, 2021.

\bibitem[\protect\citeauthoryear{Zhong \bgroup \em et al.\egroup
  }{2019}]{zhong2019graph}
Jia-Xing Zhong, Nannan Li, Weijie Kong, Shan Liu, Thomas~H Li, and Ge~Li.
\newblock Graph convolutional label noise cleaner: Train a plug-and-play action
  classifier for anomaly detection.
\newblock In {\em Proc. CVPR}, pages 1237--1246, 2019.

\bibitem[\protect\citeauthoryear{Zhu and Newsam}{2019}]{zhu2019motion}
Yi~Zhu and Shawn Newsam.
\newblock Motion-aware feature for improved video anomaly detection.
\newblock {\em arXiv preprint arXiv:1907.10211}, 2019.

\end{thebibliography}

\newpage
\appendix
\section{Tube Instance Construction}

\begin{algorithm}[h]
\caption{Tube Construction of Unary Tube Instances.}
\label{alg:1}
\begin{algorithmic}[1]
\State \textbf{Input:} Boxes set $\{B\}$, scores $S(B)$, length threshold $\zeta_1$, unary tube list $L_1=\varnothing$
    \For{$B_c$ in each category $c$}
        \While{$B_c \neq \varnothing$ }
            \State $B^p_c = \mathop{\arg\max}_{\{B_c\}} S(B_c)$, $B_{tp,m} = B^p_c$; $T$ = [$B_{tp,m}$].
            \State \emph{Forward Linking}: $t=tp$
            \While{$U(B_{t,m},B_{t+1,j}) > \lambda$, $j \in [1, N_{t+1}]$}
              \State $B_{t+1,m} = \mathop{\arg\max}_{B_{t+1,j}} S_l(B_{t,m},B_{t+1,j})$; $T$.add($B_{t+1,m}$); $t=t+1$.
            \EndWhile
            \State \emph{Backward Linking}: $t=tp$
            \While{$U(B_{t,m},B_{t-1,j}) > \lambda$, $j \in [1, N_{t-1}]$}
              \State $B_{t-1,m} = \mathop{\arg\max}_{B_{t-1,j}} S_l(B_{t,m},B_{t-1,j})$; $T$.add($B_{t-1,m}$); $t=t-1$.
            \EndWhile
            \If{len($T$) $\ge$ $\zeta_1$}:
                \State $L_1$.add($T$);
            \EndIf
                \State $\{B\}$.delete($T$).
        \EndWhile
    \EndFor
    \State \textbf{Output:} $L_1$
\end{algorithmic}
\end{algorithm}

\begin{algorithm}[h]
\caption{Tube Construction of Multivariate Tube Instances.}
\label{alg:2}
\begin{algorithmic}[1]
   \State \textbf{Input:}  Boxes set $\{B\}$, scores $S(B)$, length threshold $\zeta_2$, multivariate tube list $L_2=\varnothing$.
    \While{$B \neq \varnothing$ }
        \State  $D=\{D_t= \varnothing\}^T_{t=1}$; $B^p = \mathop{\arg\max}_{\{B\}} S(B)$; $B_{tp,m} = B^p$; $D_{tp}$ = [$B_{tp,m}$].
        \While{$U(B_{tp,m},B_{tp,j}) > \lambda$, , $j \in [1,m) \cup (m,N_{tp}]$}
            \State $D_{tp}$.add($B_{tp,j})$; $M_{tp}$ = $\cup$ ($D_{tp}$); $T$=[$M_{tp}$]. (Note: $\cup$ means union of all boxes)
        \EndWhile
        \State \emph{Forward Linking}: $t=tp$
        \While{$U(D_{t,i},B_{t+1,j}) > \lambda$, $i \in [1, N_{t}]$, $j \in [1, N_{t+1}]$}
          \State $D_{t+1}$.add($B_{t+1,j}$); $M_{t+1}$ = $\cup$ ($D_{t+1}$); $T$.add($M_{t+1}$); $t=t+1$.
        \EndWhile
        \State \emph{Backward Linking}: $t=tp$
        \While{$U(D_{t,i},B_{t-1,j}) > \lambda$, $i \in [1, N_{t}]$, $j \in [1, N_{t-1}]$}
          \State $D_{t-1}$.add($B_{t-1,j}$); $M_{t-1}$ = $\cup$ ($D_{t-1}$); $T$.add($M_{t-1}$); $t=t-1$.
        \EndWhile
        \If{len($T$) $\ge$ $\zeta_2$}:
            \State $L_2$.add($T$).
        \EndIf
        \State $\{B\}$.delete($D$).
    \EndWhile
    \State \textbf{Output:} $L_2$.
\end{algorithmic}
\end{algorithm}

To generate tube-level instance proposals in the weakly supervised manner, we first adopt Faster-RCNN \cite{ren2015faster} algorithm to detect all bounding boxes in frames, $\{B\}$, with corresponding confidence scores $S(B)$.
Subsequently, we link the detections from our single frame to produce temporarily consistent spatio-temporal tubes.
Then we illustrate the construction process of two tube-level instances in detail.

\begin{figure*}[t]
    \centering
    \includegraphics[width=0.98\linewidth]{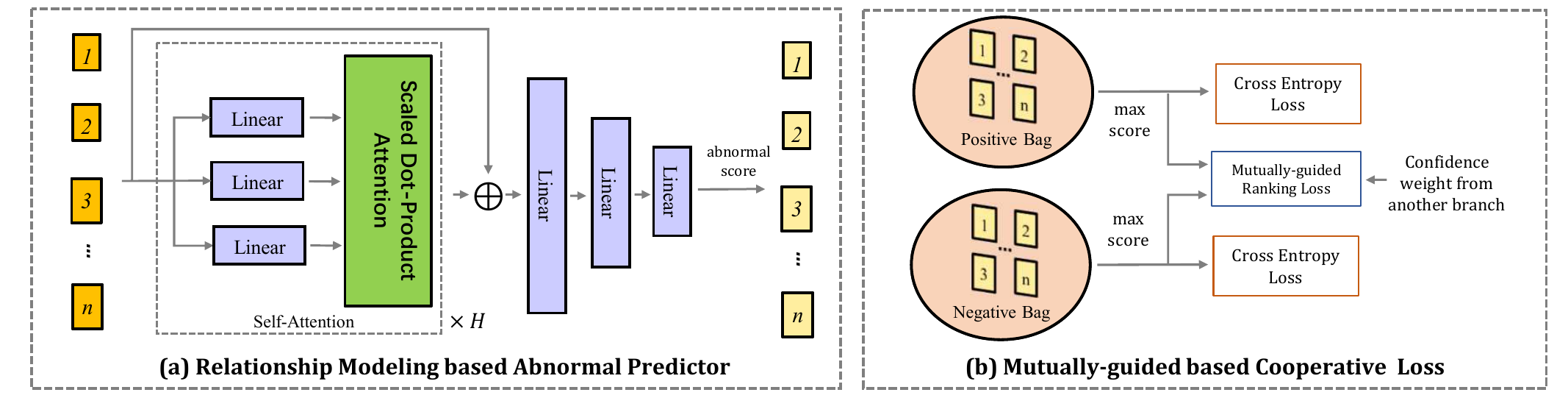}
    \caption{The illustration of relationship modeling based abnormal predictor and mutually-guided based cooperative loss.}
    \label{fig:model}
\end{figure*}

\noindent \textbf{Unary Tube}.
Unary tube encloses an individual object, which can handle some abnormal events caused by a particular object.
The tube construction process of the unary tubes can be seen as a hierarchical clustering problem.
Detection results that reflect the same object in the video are grouped into one cluster.
The tube construction process of the unary tube are outlined in detail in the Algorithm \ref{alg:1}.
For each object category $c$, we first sort the detections according to the confidence score and pick the one with the max score $B_c^p$ as the starting point of a cluster.
Then the linking process is extended both forward and backward via the greedy search algorithm and the box with the max linking score in the consecutive time is added to the corresponding cluster.
Specifically, the linking score $S_l(B_i, B_j)$ between two boxes $B_i$ and $B_j$ is defined as:
\begin{footnotesize}
\begin{equation}
\begin{split}
S_l(B_i, B_j) =
\begin{cases}
S(B_i) + S(B_j) +  \eta U(B_i, B_j) \quad \text{$I(B_i, B_j)>\lambda$}\\
0 \qquad \qquad \qquad \qquad \qquad \qquad \qquad \text{otherwise}
\end{cases},
\end{split}
\label{eq6}
\end{equation}
\end{footnotesize}
where $U(B_i, B_j)$ is the intersection-over-union (IoU) of $B_i$ and $B_j$. $\lambda$ and $\eta$ are the IoU threshold and a trade-off scalar, respectively.
We continue the linking process until there is no box could obtain the IoU greater than $\lambda$.
When a special cluster is constructed, we remove the linked boxes and collect a new cluster repeatedly until all boxes for this category are grouped.
Then we turn to next category and continue the clustering process recursively until all boxes are clustered.

\noindent \textbf{Multivariate Tube}.
Some anomalies (such as "fighting") are related to the behavior of multiple objects, so we designed the multivariate tube that contains multiple intersecting objects.
multivariate tube can capture implicit associations between objects that cannot be detected by the prior detector, modeling the relationship and interaction between multiple objects.
The construction process of the multivariate tube is outlined in the Algorithm \ref{alg:2}.
The construction of the multivariate tube also can be regarded as a clustering problem, and the condition that two boxes are grouped into the same cluster is that the IoU between the boxes is greater than $\lambda$.
Specifically, we first pick the box $B^p$ (also denoted as $B_{tp,m})$ with the max confidence score from all detections as a starting element of the cluster.
Then each other box in the same frame is determined whether it can be grouped into the current cluster $D_{tp}$.
Subsequently, the clustering process is conducted in the spatial and temporal dimensions alternately to construct a complete tube.
Spatially, each box in the next frame is compared with the clusters of the current frame $D_{t}$ to determine the cluster of the next frame $D_{t+1}$ (or $D_{t-1}$).
In the temporal dimension, the boxes of the cluster in the same frame are unionized into a joint box $M$ and follow the linking process of the unary tube to construct the tube chronologically.
After a particular tube is constructed, we remove the linked boxes in the formed cluster and continue the clustering process recursively until all boxes are clustered.

The hyper-parameters $\eta$ and $\lambda$ are empirically set to 2 and 0.1, receptively. The length threshold $\zeta_1$ and
$\zeta_2$ is 100 and 50.

\section{Framework Illustration}
We additional provide visualization illustration to show the architecture of relationship modeling based abnormal predictor and in Figure \ref{fig:model}.

\section{Training Process}
In order to better illustrate the training process of the mutually-guided progressive refinement process, the whole alternating update training algorithm is outlined in the Algorithm \ref{alg:3}.
\begin{algorithm}[h]
\caption{Mutually-guided Progressive Refinement Scheme.}
\label{alg:3}
\begin{algorithmic}[1]
\State Randomly initialize $\theta^{tube}$, $\theta^{tem}$, $\theta$ denotes the corresponding parameters.
\For {iteration=$1,...,\infty$}
 \State Randomly sample a mini-batch.
 \State \textbf{\emph{Train temporal branch:}}
 \State $\psi$ = 0, freeze $\theta^{tube}$.
   \State Obtain and compute
   $p_{t}(\mathcal{V}^{a}_{m})$, $p_{v}(\mathcal{V}^{a}_{m})$ $p_{v}(\mathcal{V}^{n}_{m})$.
    \State Compute $\mathcal{L}_{\textit{Rank}}^{\textit{tem}}$ by Equation (5).
    \State Compute $\mathcal{L}_{\textit{CE}}^{\textit{tem}} = -[ \log (p_{v}(\mathcal{V}^{a}_{m})) +  \log (1- p_{v}(\mathcal{V}^{n}_{m}))]$ .
       \State Update $\theta^{tem}$ by minimizing: $\mathcal{L} = \mathcal{L}_{\textit{CE}}^{\textit{tem}} + \mathcal{L}_{\textit{Rank}}^{\textit{tem}}$.
   \State \textbf{\emph{Train tube branch:}}
 \State $\psi$ = 1, freeze $\theta^{tem}$.
    \State Compute and obtain $p_{v}(\mathcal{T}^{a,g}_{m})$, $p_{t}(\mathcal{T}^{a,r}_{m})$ and   $p_{t}(\mathcal{T}^{n,r}_{m})$.
    \State Compute $\mathcal{L}_{\textit{Rank}}^{\textit{tube}}$ by Equation (4).
        \State Compute $\mathcal{L}_{\textit{CE}}^{\textit{tube}} = -[ \log (p_{t}(\mathcal{T}^{a,r}_{m})) +  \log (1- p_{t}(\mathcal{T}^{n,r}_{m}))]$.
    \State Update $\theta^{tube}$ by minimizing: $\mathcal{L} = \mathcal{L}_{\textit{CE}}^{\textit{tube}} + \mathcal{L}_{\textit{Rank}}^{\textit{tube}}$.
\EndFor
\end{algorithmic}
\end{algorithm}

\section{Spatio-Temporal Anomaly Detection Dataset}

\subsection{Datasets Setup}
 \begin{figure*}[t]
    \vspace{-5pt}
    \centering
    \includegraphics[width=0.90\linewidth]{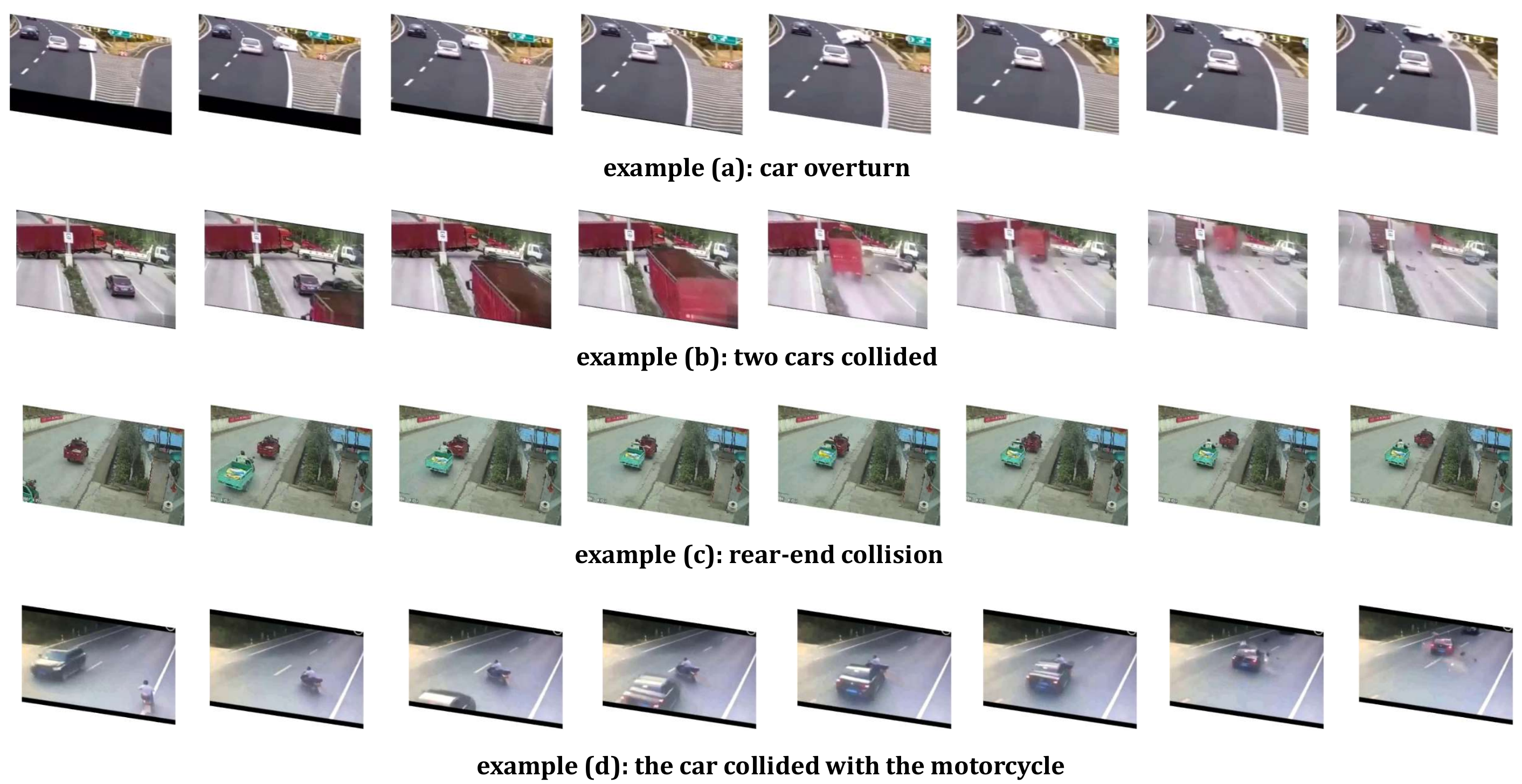}
    \caption{Examples of some anomalies from the training and testing videos in STRA.}
    \label{fig:dataset_vis}
\end{figure*}
A major challenge for the proposed WSSTAD task is the lack of appropriate datasets.
Existing datasets like ShanghaiTech \cite{luo2017revisit} and UCSD-Peds \cite{li2013anomaly} will encounter two difficulties when they are adopted for WSSTAD.
Firstly, they do not provide spatio-temporal annotations for the ground truth instances, which are necessary for evaluation.
On the other hand, the abnormalities in these datasets are relatively simple, lacking variations, and deviating from realistic situations.
To the best of our knowledge, the most suitable dataset is the Dashcam \cite{chan2016anticipating} dataset, which is designed for accident forecasting and equipped with dense bounding-box annotations. However, this dataset is too easy for the proposed task since the anomalies all occur at the end of the video.
UCF-Crime \cite{sultani2018real} is the largest anomaly detection datasets containing anomaly videos of diverse categories in complicated surveillance scenarios, which enables the model to learn various abnormal abstraction.
To this end, we build a new dataset (denoted as ST-UCF-Crime) that annotates spatio-temporal bounding boxes for abnormal events in UCF-Crime \cite{sultani2018real}.

Furthermore, \emph{traffic accident} is a common and critical category in the anomaly of surveillance video, which can be adopted to evaluate the robustness of models for anomaly detection with multiple objects interaction.
Considering that the number of traffic accidents in the ST-UCF-Crime dataset is limited, we contribute a new dataset Spatio-Temporal Road Accident (denoted as STRA) that consist of various road accidents videos on the streets, intersections or highways, such as motorcycles crash into cars, cars crash into people, etc..
STRA is collected from the traffic police department and resource-production video sharing platforms, which contributes to fine-grained anomaly detection in actual traffic accident scenarios and promoting the development of intelligent transportation. Figure \ref{fig:dataset_vis} illustrate examples of some anomalies from the training and testing videos in STRA.

\begin{figure}[h]
    \centering
    \includegraphics[width=0.98\linewidth]{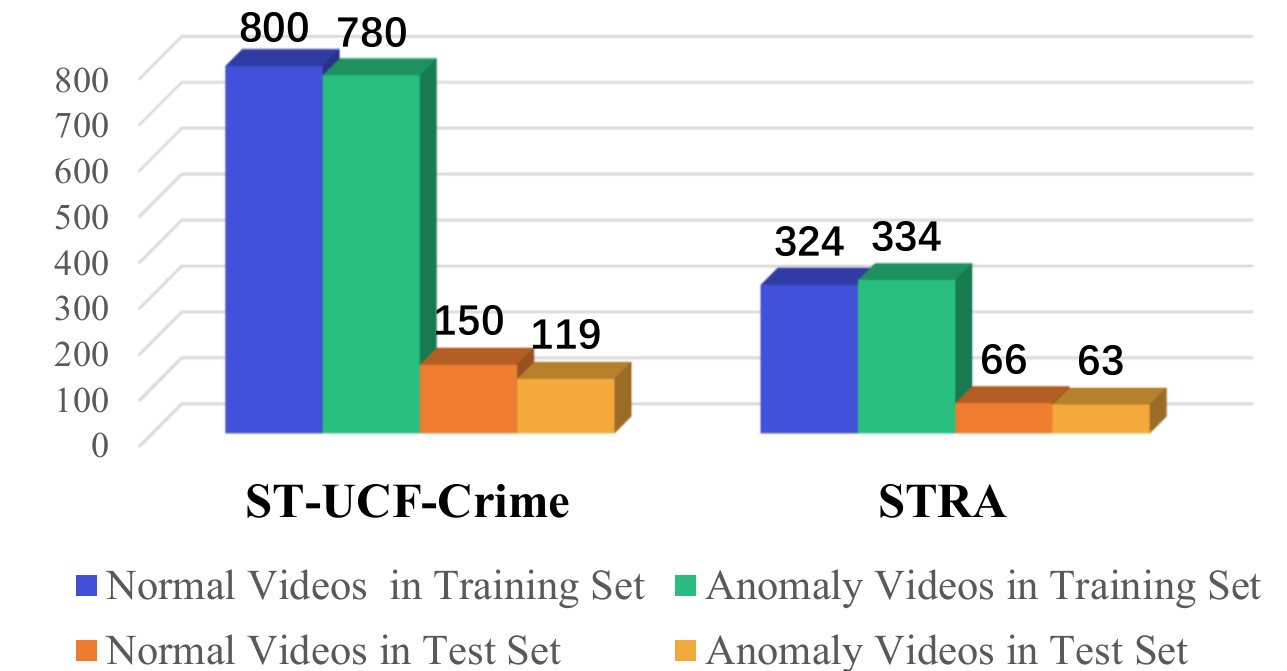}
    \caption{Video number of each split subset.}
    \label{fig:Statistics}
\end{figure}

 \begin{figure*}[t]
    \centering
    \includegraphics[width=0.88\linewidth]{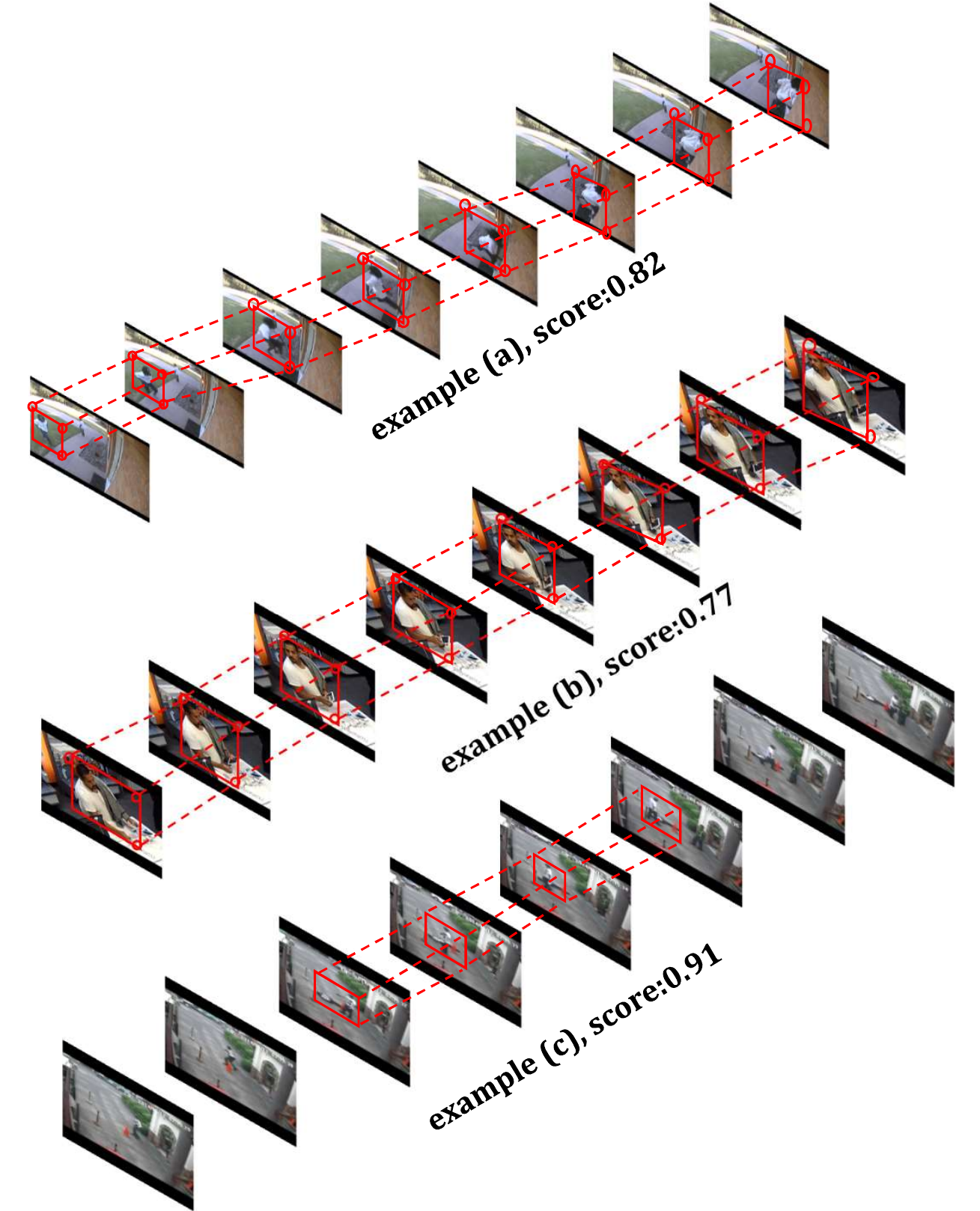}
    \caption{Visualization of a series of examples from ST-UCF-Crime with abnormal score.}
    \label{fig:example1}
\end{figure*}

\subsection{Dataset Statistics}
We remove the videos corresponds to ``explosion'' from the original UCF-Crime dataset, and obtain the reorganization of ST-UCF-Crime dataset: training set with 1580 videos and testing set with 269 videos.
For STRA, we randomly select 658 videos as the training set, and split the remaining 129 videos into the testing sets.
A group of statistics of two datasets are summarized in Figure \ref{fig:Statistics} and Table \ref{tab:dataset}.
Specifically, \emph{Average Duration} in Table \ref{tab:dataset} denotes the average duration of all videos.
\emph{Abnormal Proportion} is the proportion of abnormal frame sequences in the entire video, in which the videos are the abnormal videos of the test set.
From the table, we can observe that the abnormal moments account for a smaller percentage of the entire video in STRA, which reveals that STRA is more challenging than ST-UCF-Crime.
\emph{Annotate Box Number} indicates the number of the ground truth box annotated in each dataset.
We provide more details of two datasets in the appendix.

\begin{table}[t]
\footnotesize
\centering
\caption{Dataset statistics of ST-UCF-Crime and STRA.}
    \begin{tabular}{m{3.0cm}<{\centering}|m{2.0cm}<{\centering}|m{1.8cm}<{\centering}}
    \toprule
         & ST-UCF-Crime & STRA \\ \hline
        Video Number  & 1849	&787 \\ \hline
        Resolution  & 240 $\times$ 320	&720 $\times$ 1280 \\ \hline
        fps  & 30	&25 \\ \hline
        Average Duration  & 232.53s	&22.41s \\ \hline
        Abnormal Proportion & 16.82\%	&9.28\%\\ \hline
        Annotate Box Number& 59165	&2140\\
\bottomrule
\end{tabular}
\label{tab:dataset}
 \end{table}

\subsection{Spatio-Temporal Annotations}
In the weakly supervised setting, we only provide spatio-temporal annotations for the test set and avoid labor-intensive annotations required for the training set.
After collecting the video, we apply crowdsourced annotation based on a carefully designed annotation tool.
Each annotation worker is allowed to participate in the formal annotation work only when he/she passes the preliminary annotation test.
During the annotation task, the worker is asked to 1) watch the anomaly video; 2) annotate temporal boundaries of anomaly events (only for STRA); 3) annotate regional boundaries for the anomaly frames. Each annotations are finally reviewed by several experts for quality assurance.

\begin{table}[t]
    \footnotesize
    \centering
    \caption{False alarm rate on normal testing videos.}
     \begin{tabular}{m{3.8cm}<{\centering}|m{2.1cm}<{\centering}|m{1.2cm}<{\centering}}
    \toprule
    Baseline &   ST-UCF-Crime & STRA \\ \hline
    DMRM \cite{sultani2018real} &4.00&4.54 \\
     CIBL \cite{zhang2019temporal}&5.33 &7.58 \\
     GCLNC \cite{zhong2019graph}&3.33 &3.03 \\
     Ours &2.00 &1.52 \\
    \bottomrule
    \end{tabular}
    \label{table:result15}
     \end{table}

\section{Experiments of False Alarm Rate}
 A robust anomaly detection model should have low false alarm rate on normal videos, so we also compare the false alarm rate for some competitive methods.
The tube $\mathcal{T}_{\textit{pred}}$ of normal testing videos are selected to calculate the false alarm rate at 20\% threshold.
Table \ref{table:result15} reveals that our method can obtain a lower false alarm rate than other methods, demonstrating the robustness of our approach.
It also indicates that fine-grained spatio-temporal cues help to provide more reliable explanatory guarantees for anomaly classification.

\section{Experiments of Temporal Anomaly Detection}
 We also report the results of our method in temporal anomaly detection tasks to further prove the effectiveness of our approach. Specifically, we follow \cite{sultani2018real} to divide the whole video into equidistant sub-segments and then calculate the average scores of the dual branch. The frame-level AUC results are shown in Table \ref{table:result6}.
 We observe that our approach outperforms state-of-the-art approach \cite{sultani2018real,zhang2019temporal,zhong2019graph}, which demonstrates that our method can also competitive in temporal anomaly detection.

\begin{table}[t]
\footnotesize
\centering
\caption{Frame-level AUC comparison of some state-of-the-art methods on temporal anomaly detection.}
 \begin{tabular}{m{3.8cm}<{\centering}|m{3.0cm}<{\centering}}
\toprule
Baseline &   Frame-level AUC \\ \hline
DMRM \cite{sultani2018real} &75.41 \\
 CIBL \cite{zhang2019temporal}&78.66 \\
 GCLNC \cite{zhong2019graph}&81.08 \\
 Ours &82.73 \\
\bottomrule
\end{tabular}
\label{table:result6}
 \end{table}

 \section{Qualitative Results}

 \begin{figure*}[t]
    \centering
    \includegraphics[width=0.88\linewidth]{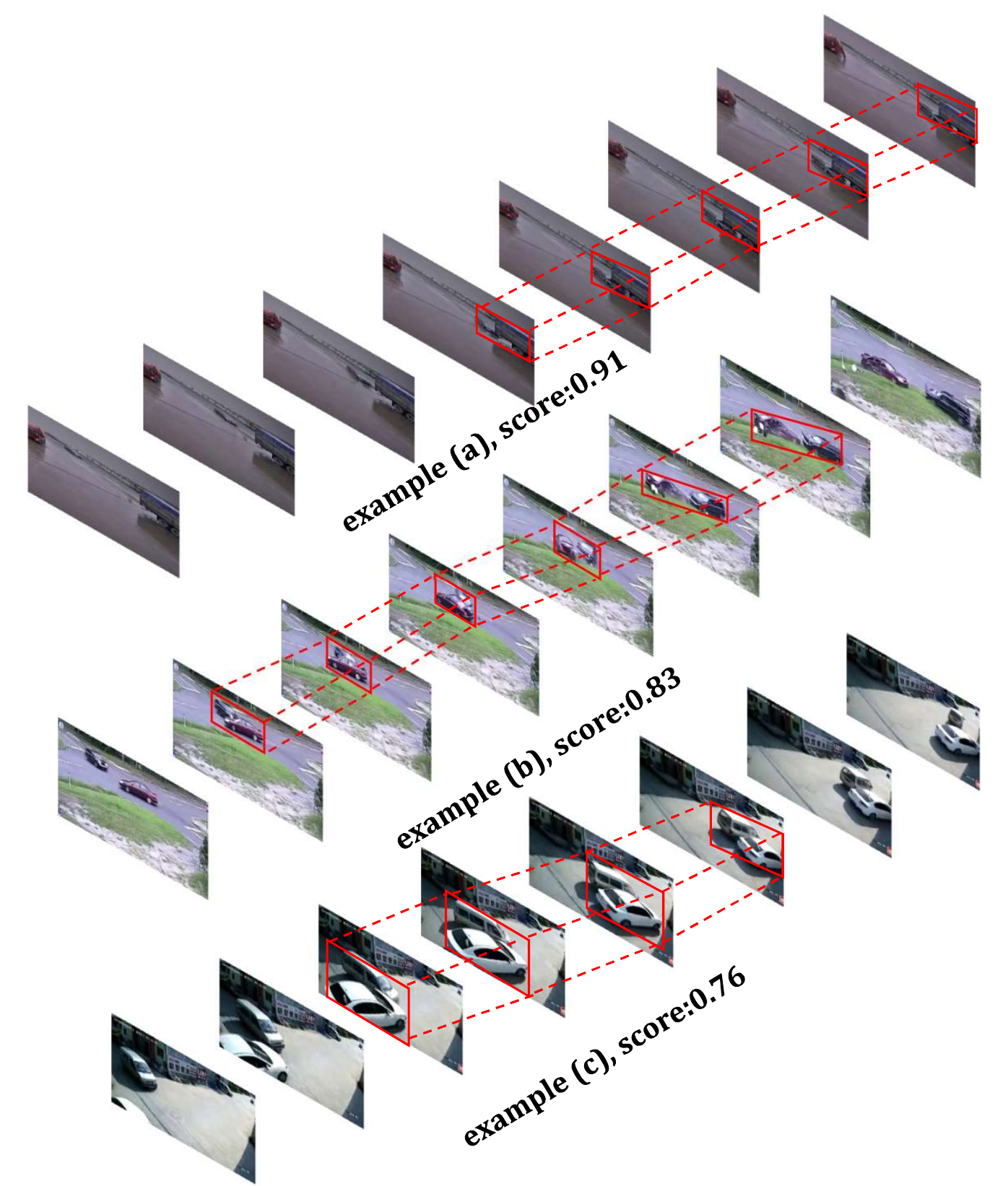}
    \caption{Visualization of a series of examples from STRA with abnormal score.}
    \label{fig:example2}
\end{figure*}

Figure \ref{fig:example1} and Figure \ref{fig:example2} show six prediction examples and the abnormal scores provided by our approach.
The examples in Figure \ref{fig:example1} are from
 ST-UCF-Crime, and the examples in Figure \ref{fig:example2} are from STRA.
From the figures, we have the following observations:
1) our approach can infer the abnormal tube accurately with high abnormal scores.
2) our approach helps to predict different kinds of anomalies accurately, such as ``burglary''  in example (a), ``shoplifting'' in example (b), and ``shooting'' in example (c).
3) our approach can locate the corresponding tube for traffic accidents in different road conditions.
The above observations illustrate the robustness of our approach under different kinds of anomalies in different scenarios.

 \begin{figure*}[t]
    \centering
    \includegraphics[width=0.88\linewidth]{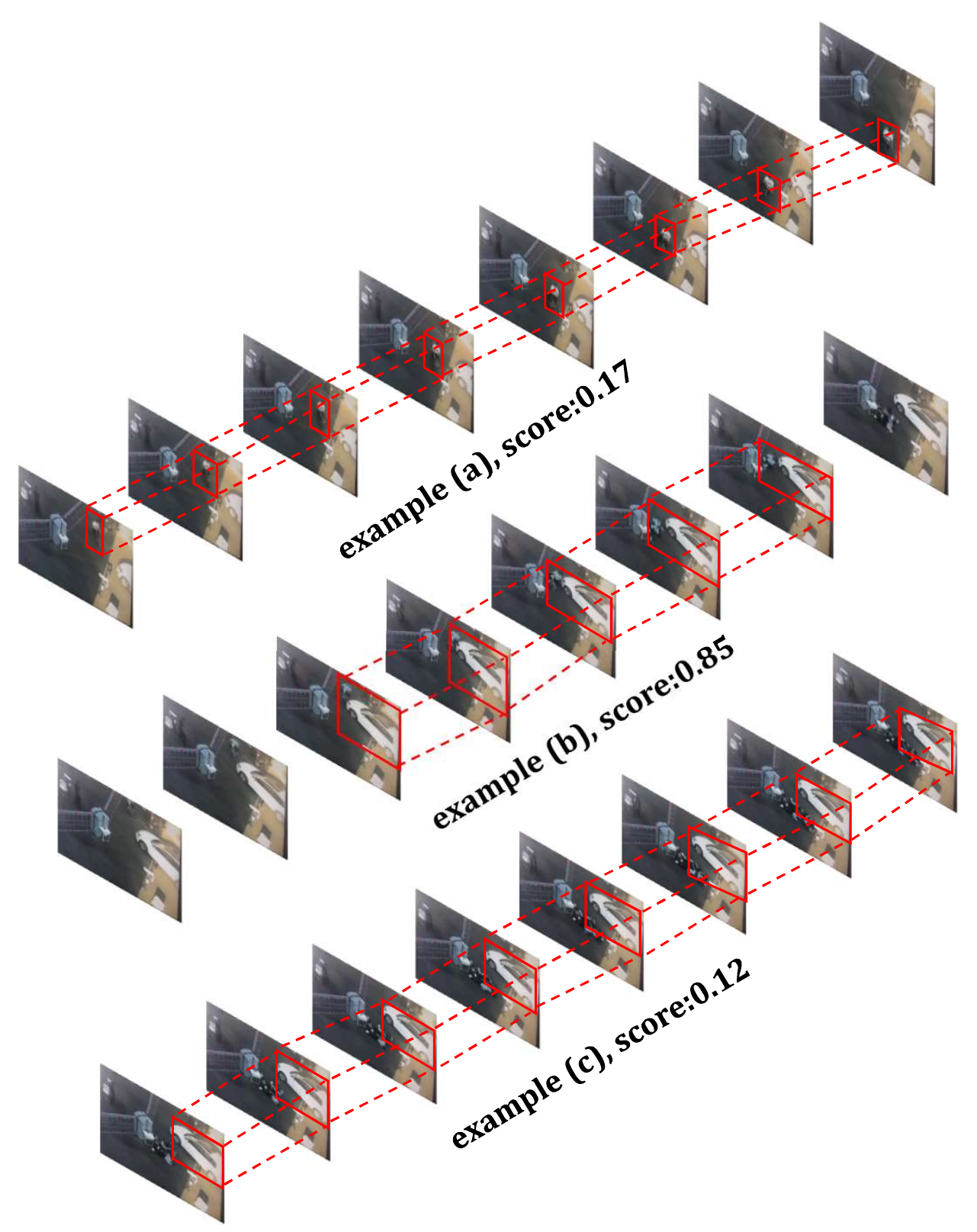}
    \caption{Visualization of tube instances and their abnormal scores from abnormal videos.}
    \label{fig:example3_1}
\end{figure*}

 \begin{figure*}[t]
    \centering
    \includegraphics[width=0.88\linewidth]{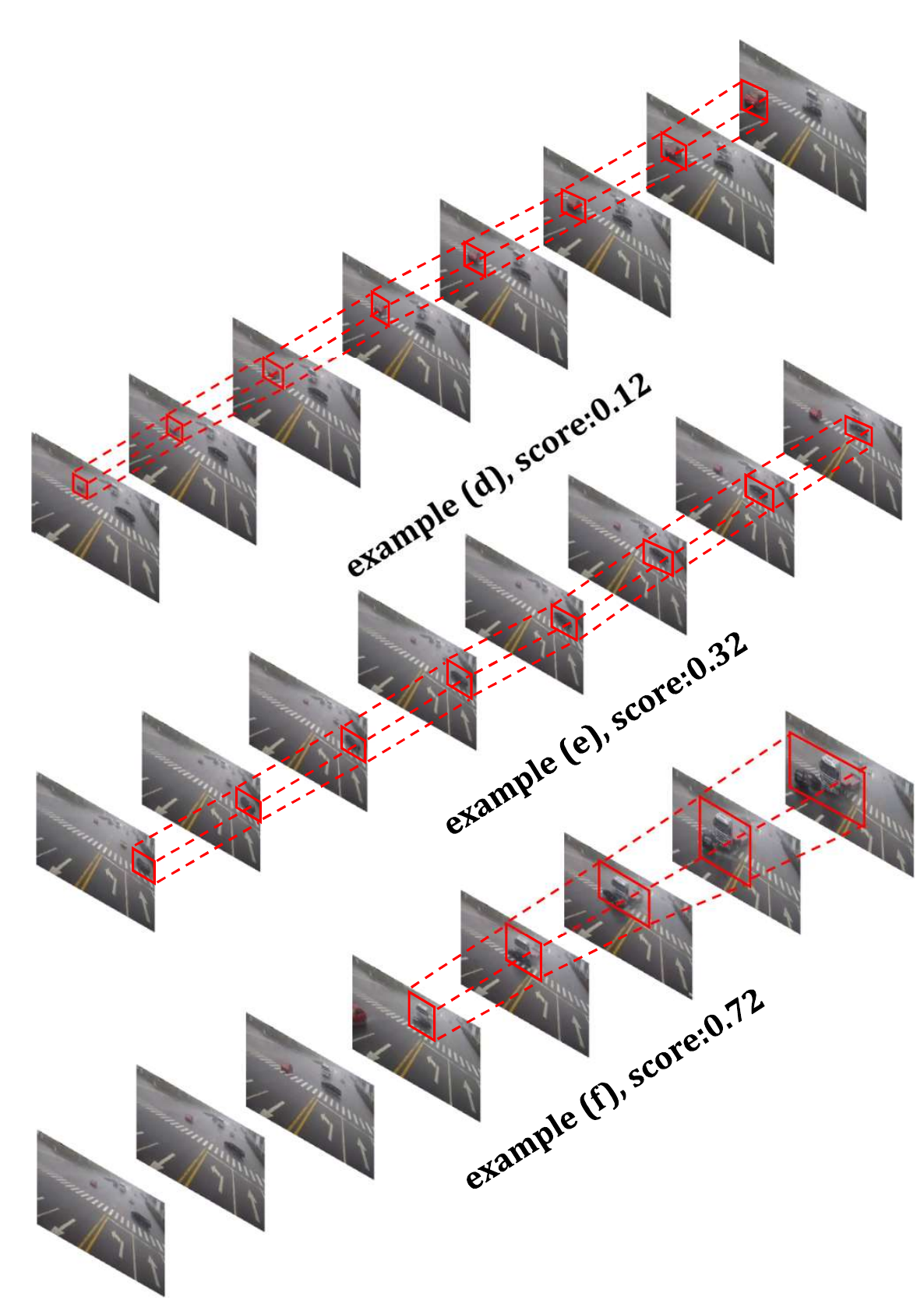}
    \caption{Visualization of tube instances and their abnormal scores from abnormal videos.}
    \label{fig:example3_2}
\end{figure*}

Furthermore, to investigate the model's performance on specific anomalous videos, we randomly select two videos from the test set of STRA and predict abnormal scores for six of the tube instances. The results are shown in Figure \ref{fig:example3_1} and \ref{fig:example3_2}.
As depicted in Figure \ref{fig:example3_1} example(b), our approach can accurately assign a high abnormal score (i.e., 0.85) for the accident event of the car hit the motorcycle. For the scene of normal motorcycle driving (example (a)) and vehicle parking (example (c)), our model gives a lower anomaly score.
It reveals that our model tends to make accurate abnormal predictions for different actual road conditions.

 \begin{figure*}[t]
    \centering
    \includegraphics[width=0.90\linewidth]{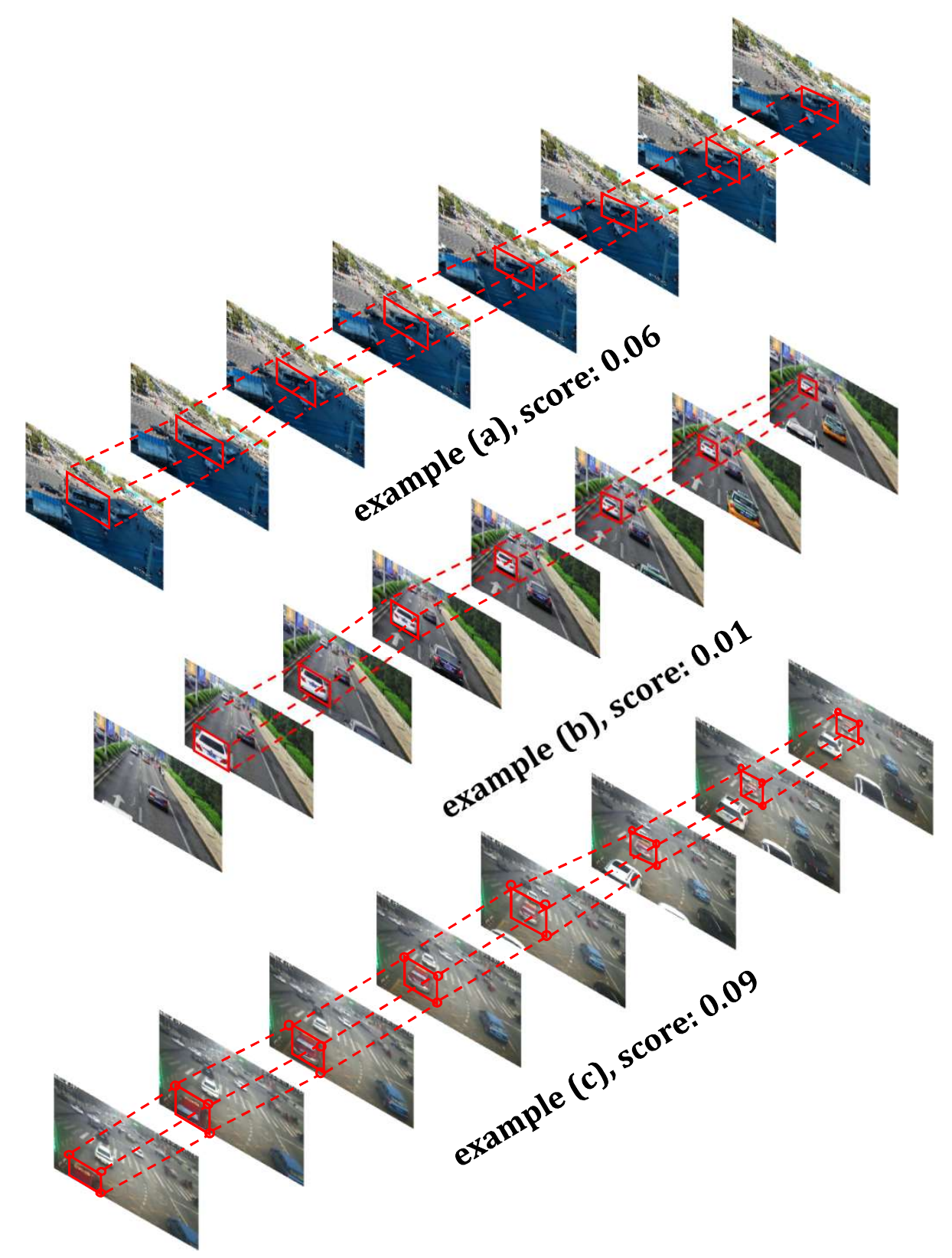}
    \caption{Visualization of tube instances and their abnormal scores from normal videos.}
    \label{fig:example4}
\end{figure*}

To qualitatively analyze the false alarm performance of our approach, we randomly choose three normal video from the test set in STRA, and visualize the max instance and their scores. The results are shown in the Figure 7. As shown in Figure \ref{fig:example4}, our method attempt to provide low abnormal scores for normal traffic scenes, which indicates that our approach is also robust for normal videos.

\end{document}